\documentclass[wcp]{jmlr}

\usepackage{longtable}

\usepackage{booktabs}

\usepackage{booktabs}
\usepackage{comment}

\usepackage{lineno}

\makeatletter
\let\Ginclude@graphics\@org@Ginclude@graphics 
\makeatother

\jmlrvolume{222}
\jmlryear{2023}
\jmlrworkshop{ACML 2023}

\title[FedPPD]{Federated Learning with Uncertainty via Distilled \titlebreak Predictive Distributions}

 \author{\Name{Shrey Bhatt$^*$} \Email{shreyb1997@gmail.com}\\
 \addr Eightfold AI
 \AND
\Name{Aishwarya Gupta$^*$} \Email{aishwaryag@cse.iitk.ac.in} \\
\addr IIT Kanpur
\AND
\Name{Piyush Rai} \Email{piyush@cse.iitk.ac.in}\\
\addr IIT Kanpur
}

\editors{Berrin Yan{\i}ko\u{g}lu and Wray Buntine}

\begin{document}

\maketitle
\def\thefootnote{*}\footnotetext{Equal contribution. A significant portion of the work was done when Shrey Bhatt was a Masters student at IIT Kanpur.}

\begin{abstract}
Most existing federated learning methods are unable to estimate model/predictive uncertainty since the client models are trained using the standard loss function minimization approach which ignores such uncertainties. In many situations, however, especially in limited data settings, it is beneficial to take into account the uncertainty in the model parameters at each client as it leads to more accurate predictions and also because reliable estimates of uncertainty can be used for tasks, such as out-of-distribution (OOD) detection, and sequential decision-making tasks, such as active learning. We present a framework for federated learning with uncertainty where, in each round, each client infers the posterior distribution over its parameters as well as the posterior predictive distribution (PPD), distills the PPD into a single deep neural network, and sends this network to the server. Unlike some of the recent Bayesian approaches to federated learning, our approach does not require sending the whole posterior distribution of the parameters from each client to the server but only the PPD in the distilled form as a deep neural network. In addition, when making predictions at test time, it does not require computationally expensive Monte-Carlo averaging over the posterior distribution because our approach always maintains the PPD in the form of a single deep neural network. Moreover, our approach does not make any restrictive assumptions, such as the form of the clients' posterior distributions, or of their PPDs. We evaluate our approach on classification in federated setting, as well as active learning and OOD detection in federated settings, on which our approach outperforms various existing federated learning baselines.
\end{abstract}

\section{Introduction}

Federated learning~\citep{kairouz2021advances} enables collaborative learning from distributed data located at multiple clients without the need to share the data among the different clients or with a central server. Much progress has been made in recent work on various aspects of this problem setting, such as improved optimization at each client~\citep{li2020federatedheterogenous}, improved aggregation of client models at the server~\citep{chen2020fedbe}, handling the heterogeneity in clients' data distributions~\citep{zhu2021data}, and also efforts towards personalization of the client models~\citep{mansour2020three}.

Most existing formulations of federated learning view it as an optimization problem where the global loss function is optimized over multiple rounds, with each round consisting of point estimation of a loss function defined over the client's local data, followed by an aggregation of client models on a central server. Point estimation, however, is prone to overfitting especially if the amount of training data on clients is very small. Moreover, crucially, such an approach ignores the uncertainty in the client models. Indeed, taking into account the model uncertainty has been shown to be useful not just for improved accuracy and robustness of predictions when the amount of training data is limited, as well as in other tasks, such as out-of-distribution (OOD) detection~\citep{salehi2021unified} and active learning~\citep{ahn2022federated}. In this work, we present a probabilistic approach to federated learning which takes into account the model uncertainty at each client (by learning a \emph{posterior distribution}, i.e., the conditional distribution of the model parameters given the training data), and also demonstrate its effectiveness for other tasks in federated settings where accurate estimates of model uncertainty are crucial, such as OOD detection and active learning in federated setting.

Despite its importance, federated learning in the setting where each client learns a posterior distribution is inherently a challenging problem. Unlike standard federated learning, in this setting, each client needs to estimate the posterior distribution over its weights using Bayesian inference, and also the posterior predictive distribution (PPD) which needed at the prediction stage, which is an intractable problem.
Typical ways to address this intractability of Bayesian inference for deep learning models include (1) Approximate Bayesian inference where the posterior distribution of model parameters is usually estimated via approximate inference methods, such as MCMC~\citep{zhang2019cyclical,izmailov2021bayesian}, variational inference~\citep{zhang2018advances}, or other faster approximations such as modeling the posterior via a Gaussian distribution constructed using the SGD iterates~\citep{maddox2019simple}, or (2) ensemble methods, such as deep ensembles~\citep{lakshminarayanan2017simple} where the model is trained using different initialization to yield an ensemble whose diversity represents the model uncertainty.

The other key challenge for federated learning in this setting is efficiently communicating the client model parameters, which are represented by a probability distribution, to the server, and their aggregation at the server. Note that, unlike standard federated learning, in our setting, to capture client model's uncertainty, each client needs to maintain either a probability distribution over its model weights or an ensemble over the model weights. Both of these approaches make it difficult to efficiently communicate the client models and aggregate them at the server. Some recent attempts towards such settings of federated learning have relied on simplifications such as assuming that the posterior distribution of each client's weights is a Gaussian~\citep{al2020federated,linsner2021approaches}, which makes model communication and aggregation at the server somewhat easier. However, this severely restricts the expressiveness of the client models. In our work, we do not make any assumption on the form of the posterior distribution of the client weights. Another appealing aspects of our federated learning approach is that, at test time, it does not require Monte-Carlo averaging~\citep{bishop2006pattern,korattikara2015bayesian} which is usually required by Bayesian methods (especially for non-conjugate models, such as deep learning models) at test time, making them slow (essentially, using $m$ Monte-Carlo samples from the posterior makes prediction $m$ times slower). In contrast, our approach leverages ideas from the distillation of posterior predictive distribution (PPD)~\citep{korattikara2015bayesian}, using which we are able to represent the entire posterior predictive distribution using a single deep neural network, resulting in fast predictions at test time.

Our contributions are summarized below
\begin{itemize}
    \item We present a novel and efficient probabilistic framework to federated learning in which each client performs a distillation of its posterior predictive distribution into a single deep neural network. This allows solving the problem of federated learning with client model uncertainty using ideas developed for standard federated learning methods, while still capturing and leveraging model uncertainty. 
    \item Our approach does not make any strict assumptions on the form of the clients' posterior distributions (e.g., Gaussian~\citep{al2020federated}) or predictive distributions. Moreover, despite each client learning a posterior distribution, our approach is still fast at test time since it does not require Monte-Carlo averaging (which is akin to averaging over an ensemble) but uses the idea of distribution distillation to represent the posterior predictive distribution (PPD) via a single deep neural network.
    \item We present various ways to aggregate the clients' predictive distributions at the server, both with as well as without requiring publicly available (unlabeled) data at the server. 
    \item In addition to tasks such as classification and out-of-distribution (OOD) detection, we also show a use case of our approach for the problem of active learning in federated setting~\citep{ahn2022federated} where our approach outperforms existing methods.
\end{itemize}

\section{Bayesian Federated Learning via Predictive Distribution Distillation}
\begin{figure*}[t]
    \centering
    \includegraphics[scale=0.4]{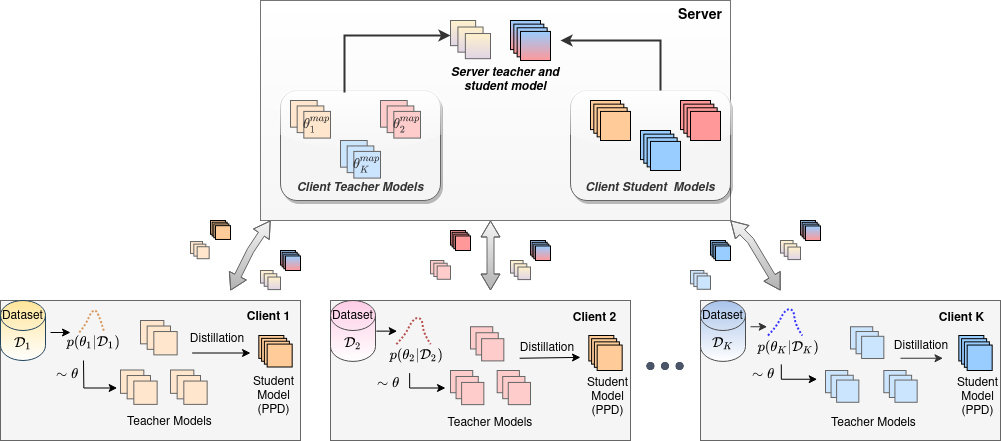}
    \caption{\small{The above figure summarizes our framework. Each client infers the (approximate) posterior distribution by generating the posterior samples (teacher models) which are distilled to give the PPD (student model parameterized by a deep neural network). Each client communicates its MAP teacher sample and the PPD to the server which aggregates them to yield a global teacher sample and the global PPD, both of which are sent back to the clients, which use these quantities for the next round of learning.}}
    \label{fig:fedppd_diagram}
    \vspace{-1em}
\end{figure*}

Unlike standard federated learning where the client model is represented by a single neural network whose weights are estimated by minimizing a loss function using client's data, we consider the setting of federated learning where each client learns a posterior distribution over its weights. The posterior distribution $p(\theta | \mathcal{D})$ is a probability distribution of network weights representing how likely a sample $\theta$ explains the training data $\mathcal{D}$. The goal is to efficiently communicate the clients' local posteriors to the server and aggregate these local posteriors to learn a global model that can serve all the clients.

However, since we usually care about predictive tasks, the actual quantity of interest in our probabilistic setting is not the posterior distribution per se, but the posterior predictive distribution (PPD). Given a set of $m$ samples $\theta^{(1)},\ldots,\theta^{(m)}$ from the posterior, estimated using some training data $\mathcal{D}$, the Monte Carlo approximation of the PPD of of a test input $x$ is defined as $p(y|x,\mathcal{D}) = \frac{1}{m}\sum_{i=1}^m p(y|x,\theta^{(i)})$. Note that the PPD can be thought of as an ensemble of $m$ models drawn i.i.d. from the posterior.

Since the PPD is the actual quantity of interest, in our probabilistic federated learning setting, we aim to directly estimate the PPD at each client. However, even estimating and representing the PPD has challenges. In particular, since the PPD is essentially an ensemble of models, storing and communicating such an ensemble from each client to the server can be challenging. To address this issue, we leverage the idea of distribution/ensemble distillation~\citep{korattikara2015bayesian}, where the PPD of a deep learning model can be efficiently distilled and stored as a single deep neural network. We leverage this distillation idea on each client to represent the client's PPD using a single neural network which can then be communicated and aggregated at the server in pretty much the same way as it is done in standard federated learning. We also note that, although we use the distillation framework proposed in~\citep{korattikara2015bayesian}, our approach is general and can leverage various other recently proposed methods for distribution/ensemble distillation.

Our approach can be summarized as follows (and is illustrated in Figure.~\ref{fig:fedppd_diagram})

\begin{enumerate}
    \item For each client, we perform approximate Bayesian inference for the posterior distribution of the client model weights using Markov Chain Monte Carlo (MCMC) sampling. This gives us a set of samples from the client's posterior and these samples will be used as teacher models which we distill into a student model. We use stochastic gradient Langevin dynamics (SGLD) sampling~\citep{welling2011bayesian} since it gives us an online method to efficiently distill these posterior samples into a student model (step 2 below).
    \item For each client, we distill the MCMC samples (teacher models) directly into the posterior predictive distribution (PPD), which is the student model. Notably, in this distillation based approach~\citep{korattikara2015bayesian}, the PPD for each client is represented succinctly by a \emph{single} deep neural network, instead of via an ensemble of deep neural network. This makes prediction stage much faster as compared to typical Bayesian approaches. 
    \item For each client, the teacher model with largest posterior probability (i.e., the MAP sample) from its posterior distribution and the student model representing the client's PPD (both of which are deep neural networks), are sent to the server. 
    \item The server aggregates the teacher and student models it receives from all the clients. For the aggregation, we consider several approaches described in Sec~\ref{sec:aggr}.
    \item The aggregated teacher and student models are sent back to each client, and the process continues for the next round.
    \item We continue steps 1-5 till convergence.
\end{enumerate}

\subsection{Posterior Inference and Distillation of Client's PPD}
\label{sec:fedbdk-1}
We assume there are $K$ clients with labeled data $\mathcal{D}_1,\ldots,\mathcal{D}_K$, respectively. On each client, we take the Monte Carlo approximation of its posterior predictive distribution (PPD) and distill it into a single deep neural network using an online Bayesian inference algorithm, as done by the Bayesian Dark Knowledge (BDK) approach in~\citep{korattikara2015bayesian}. Each iteration of this distillation procedure first generates a sample from the client's posterior distribution using the stochastic Langevin gradient descent (SGLD) algorithm~\citep{welling2011bayesian} and incrementally ``injects'' the sample into a deep neural network $\mathcal{S}$ (referred to as ``student'') with parameters $w$, representing a succinct form of the client's (approximate) PPD. This is illustrated by each of the client blocks shown in Figure~\ref{fig:fedppd_diagram}. For client $k$, assuming the set of samples generated by SGLD to be $\theta_k^{(1)},\ldots,\theta_k^{(m)}$, this distillation procedure can be seen as learning the parameters $w_k$ of the client $k$'s student model $\mathcal{S}_k$ by minimizing the following loss function~\citep{korattikara2015bayesian} using an unlabeled distillation dataset $\mathcal{D}_k^\prime$ at client $k$

\begin{equation}
    \hat{L}(w_k) = - \frac{1}{m} \sum_{i=1}^m \sum_{x^{'} \in \mathcal{D}^{'}_k} \mathbb{E}_{p(y=j | x', \theta_k^{(i)})} \log \mathcal{S}_k (y=j | x', w_k)
\end{equation}

Note that, in the above equation, $\log \mathcal{S}_k (y=j | x', w_k)$ is the log of the student model output value indicating the predicted probability of the label $y$ taking value $j$ for some input $x'$. The unlabeled dataset $\mathcal{D}_k^\prime$ can be generated from the original labeled dataset $\mathcal{D}_k$ by adding perturbations to the inputs, as suggested in~\citep{korattikara2015bayesian}. 

We sketch the full algorithm for optimizing for $w_k$ in the appendix. We use this algorithm at each client to learn the student model $\mathcal{S}_k$ which represents a compact approximation of the client $k$'s PPD in form of a single deep neural network (as shown in the client block in Figure~\ref{fig:fedppd_diagram}), which can be now communicated to the server just like client models are communicated in standard federated learning algorithms. Note that, as shown in Figure~\ref{fig:fedppd_diagram}, in our federated setting, in addition to weights $w_k$ of its PPD approximation (the student model), each client $k$ also sends the approximate maximum-a-posteriori (MAP) sample $\theta_k^{MAP}$ defined as the sample $\theta_k^{(i)}$, $i \in \{1,2,\ldots,m\}$ with the largest posterior density $p(\theta_k^{(i)}|\mathcal{D}_k)$. Note that $\theta_k^{MAP}$ is typically an \emph{approximate} MAP sample since the posterior $p(\theta_k|\mathcal{D})$ itself is approximated using sampling. The overall sketch of our federated learning procedure, which we call
FedPPD (Federated Learning via Posterior Predictive Distributions), is shown in Algorithm~\ref{algo}.

\begin{algorithm2e}
\caption{FedPPD}
\label{algo}
\DontPrintSemicolon
\KwData{Number of communication rounds $T$, total clients $K$, unlabeled dataset $\mathcal{U}  = \{x_i\}_{i=1}^P$, server teacher model weights $\theta_g$, server student model weights $w_g$, client teacher model weights $\{\theta_i\}_{i=1}^{K}$, client student model weights $\{w_i\}_{i=1}^K$, number of training samples at client $\{n_i\}_{i=1}^K$}
\BlankLine
\KwResult{Final Server Student Model Weight $w_g^{(T)}$}
\BlankLine
\For{each round $t = 0, \ldots, T-1$}{
    Server broadcasts $\theta_g^{(t)}$ and $w_g^{(t)}$\;
    \BlankLine
    \For{each client $i \in \{1, \dots, K\}$}{
        $\theta_i   = \theta_g^{(t)}$, $w_i = w_g^{(t)}$\;
        Update $\theta_i$ and $w_i$ locally as per \citep{korattikara2015bayesian}\;
    }
    Communicate $\{\theta_i^{MAP}\}_{i=1}^K$ and $\{w_i\}_{i=1}^K$ to server\;
    $\theta_g^{(t+1)}$, $w_g^{(t+1)}$ = Server\_Update($\{\theta_i^{MAP}\}_{i=1}^K,  \{w_i\}_{i=1}^K, \{n_i\}_{i=1}^K$)\;
}
\end{algorithm2e}

\subsection{Aggregation of Client Models}
\label{sec:aggr}

As described in the previous section, the server receives two models from client $k$ - the (approximate) MAP sample $\theta_k^{MAP}$ (the teacher) as well as the (approximate) PPD $w_k$ (the student). We denote the teacher models (approximate MAP samples) from the $K$ clients as $\{\theta_1^{MAP},\ldots,\theta_K^{MAP}\}$ and the respective student models (approximate PPD) as $\{w_1,\ldots,w_K\}$. These models are aggregated at the server and then sent back to each client for the next round. We denote the server aggregated quantities for the teacher and student models as $\theta_g$ and $w_g$ (we use $g$ to refer to ``global'').

In this work, we consider and experiment with two aggregation schemes on the server.
\textbf{Simple Aggregation of Client Models:} Our first aggregation scheme (shown in Algorithm~\ref{algo-agg}) computes dataset-size-weighted averages of all the teacher models and all the student models received at the server. Denoting the number of training examples at client $k$ as $n_k$ and $N = \sum_{k=1}^K n_k$, we compute $\theta_g = \frac{1}{N}\sum_{k=1}^K n_k\theta_k^{MAP}$ and $w_g = \frac{1}{N}\sum_{k=1}^K n_k w_k$, similar to how FedAvg algorithm~\citep{mcmahan2017communication} aggregates client models on the server.

\begin{algorithm2e}
\caption{Server\_Update (Average)}
\label{algo-agg}
\DontPrintSemicolon
\BlankLine
\KwData{client teacher model's MAP estimate $\{\theta_i^{MAP}\}_{i=1}^{K}$, client student model weights $\{w_i\}_{i=1}^K$, number of training samples at client $\{n_i\}_{i=1}^K$}
\BlankLine
\KwResult{Resultant Teacher Model $\overline{\theta}$, Student Model $\overline{w}$}
\BlankLine
$N = \sum_{i=1}^K n_i$ \tcc*[r]{total number of samples}\;
$\overline{\theta} = \frac{1}{N}\sum_{i=1}^K n_i \theta_i^{MAP}$\;
\BlankLine
$\overline{w} = \frac{1}{N}\sum_{i=1}^K n_i w_i$\;
\end{algorithm2e}

\textbf{Distillation-based Aggregation of Client Models:} Our second aggregation scheme goes beyond computing (weighted) averages of models received only from the clients. The motivation behind this approach is that the client models (both teachers as well as students) received at the server may not be diverse enough to capture the diversity and heterogeneity of the clients~\citep{chen2020fedbe}. To address this issue, this approach (shown in Algorithm~\ref{algo-distill})  first fits two probability distributions, one over the $K$ teacher models and the other over the $K$ student models received from the clients. It then uses these distributions to generate $M$ \emph{additional} client-\emph{like} teacher models and student models. Using the actual teacher models (resp. student models) and the additionally \emph{generated} teacher models (resp. student models), we perform knowledge distillation on the server to compute the global teacher model $\theta_g$ and the global student model $w_g$. This server-side distillation procedure requires an \emph{unlabeled} dataset $\mathcal{U}$ on the server. Applying the actual and generated teacher models (resp. student models) on the unlabeled dataset $\mathcal{U}$ gives us pseudo-labeled data $\mathcal{T}$ where each pseudo-label is defined as the averaged prediction (softmax probability vector) obtained by applying the actual and generated teacher models (resp. student models) to an unlabeled input. For the distillation step, we finally run the Stochastic Weighted Averaging (SWA) algorithm~\citep{izmailov2018averaging} using the pseudo-labeled data $\mathcal{T}$ and the simple aggregation of the client models as initialization. Both $\theta_g$ and $w_g$ can be obtained by following this procedure in an identical manner. Recently, this idea was also used in Federated Bayesian Ensemble (FedBE)~\citep{chen2020fedbe} to learn the global model.

The two aggregation schemes for server-side updates are shown in Algorithm~\ref{algo-agg}, and \ref{algo-distill}. Note that, among the two aggregation schemes, 
only Algorithm~\ref{algo-distill} assumes the availability of unlabeled dataset at the server.
Also, owing to high computation capacity, server can compute $\theta_g$ and $w_g$ in parallel for all the aggregation schemes; incurring no additional delays in communication rounds.

\begin{algorithm2e}
\caption{Server\_Update (Distill)}
\label{algo-distill}
\DontPrintSemicolon
\BlankLine
\KwData{Unlabeled dataset $\mathcal{U}$, client teacher model's MAP estimate $\{\theta_i^{MAP}\}_{i=1}^{K}$, client student model weights $\{w_i\}_{i=1}^K$, number of training samples at client $\{n_i\}_{i=1}^K$}
\BlankLine
\KwResult{Resultant Teacher Model $\theta_g$, Student Model $w_g$}
\BlankLine
$N = \sum_{i=1}^K n_i$ \tcc*[r]{total number of samples}\;
$\overline{\theta} = \frac{1}{N}\sum_{i=1}^K n_i \theta_i^{MAP}$, \quad $\overline{w} = \frac{1}{N}\sum_{i=1}^K n_i w_i$\;
\BlankLine
\Begin{
Construct global teacher model distribution $p(\theta | D)$ from $\{\theta_i^{MAP}\}_{i=1}^K$ \tcc*[r]{using Gaussian approximate}\;
Sample $M$ additional teachers and form teacher ensemble \;
$E_T=\{\theta_j \sim p(\theta | \mathcal{D})\}_{j=1}^{M} \cup \{\overline{\theta}\} \cup \{\theta_i\}_{i=1}^{K}$ \;
\BlankLine
Annotate $\mathcal{U}$ using $E_T$ to generate pseudo-labeled dataset $\mathcal{T}$ \;
Distill $E_T$ knowledge to $\overline{\theta}$ using SWA : $\theta_g = SWA(\overline{\theta}, E_T, \mathcal{T})$ \;
}
\BlankLine
Similarly follow the above steps with $\{w_i\}_{i=1}^K$ and $\overline{w}$ to get $w_g$ \;
\end{algorithm2e}

\section{Related Work}

Federated learning has received considerable research interest recently. The area is vast and we refer the reader to excellent surveys~\citep{li2020federated,kairouz2021advances} on the topic for a more detailed overview. In this section, we discuss the works that are the most relevant to our work.

While standard federated learning approaches assume that each client does point estimation of its model weights by optimizing a loss function over its own data, recent work has considered posing federated learning as a posterior inference problem where a global posterior distribution is inferred by aggregating local posteriors computed at each client. FedPA~\citep{al2020federated} is one such recent approach which performs approximate inference for the posterior distribution of each client's weights. However, it assumes a restrictive form for the posterior (Gaussian), as also assumed in some other recent works~\citep{liu2021bayesian,guo2023federated}. Moreover, the method needs to estimate the covariance matrix of the Gaussian posterior, which is difficult in general and approximations are needed. Moreover, although FedPA estimates the (approximate) posterior on each client, due to efficiency/communication concerns, at the server, it only computes a point estimate (the mean) of the global posterior. Thus, even though the approach is motivated from a Bayesian setting, in the end, it does not provide a posterior distribution or a PPD for the global model.

Recently, \citep{linsner2021approaches} presented methods for uncertainty quantification in federated learning using a variety of posterior approximation methods for deep neural networks, such as Monte Carlo dropout~\citep{gal2016dropout}, stochastic weight averaging Gaussian (SWAG)~\citep{maddox2019simple}, and deep ensembles~\citep{lakshminarayanan2017simple}. These approaches, however, also suffer from poor quality of approximation of the posterior at each client. \citep{lee2020bayesian} also propose a Bayesian approach for federated learning. However, their approach also makes restrictive assumptions, such as the distribution of the gradients at each of the clients being jointly Gaussian.

Probabilistic/Bayesian approaches for federated learning have also been proposed in recent work in the context of learning \emph{personalized} models at each client, using ideas such as Gaussian Process~\cite{achituve2021personalized} and variation inference~\cite{zhang2022personalized}. In contrast to these works, our setting is aimed at learning a global model that can be served to all the clients.

Instead of a simple aggregation of client models at the server, FedBE~\citep{chen2020fedbe} uses the client models to construct a distribution at the server and further distills this distribution into a single model using distillation.
However, FedBE only performs point estimation ignoring any uncertainty in the client models. Another probabilistic approach to federated learning \cite{thorgeirsson2020probabilistic} fits a Gaussian distribution using the client models, and sends the mean of this Gaussian to each client for the next round of client model training. This approach too does not estimate a posterior at each client, and thus ignores the uncertainty in client models. 

In the context of Bayesian learning, recent work has also explored federated versions of Markov Chain Monte Carlo sampling algorithms, such as stochastic gradient Langevin dynamics sampling~\citep{lee2020bayesian,el2021federated}. While interesting in their own right in terms of performing MCMC sampling in federated settings, these methods are not designed with the goal of real-world applications of federated learning, where fast prediction and compact model sizes are essential.

Among other probabilistic approaches to federated learning, recent work has explored the use of latent variables in federated learning. In \cite{louizos2021expectation}, a hierarchical prior is used on client model's weights where the prior's mean is set to the server's global model, and additional latent variables can also be used to impose other structures, such as sparsity of client model weights. However, these approaches do not model the uncertainty in the client model.

Some of the recent work on federated learning using knowledge distillation is also relevant. Note that our work leverages the ideas of teacher-student distillation, both at the clients (when learning a representation of the PPD using a single deep neural network), as well as in our second aggregation strategy where server-side distillation is used for learning the global model. In federated learning, the idea of distillation has been used in other works as well, such as federated learning when the client models are of different sizes and the (weighted) averaging does not make sense due to the different size/architecture of different client models~\citep{zhu2021data}. Moreover, server-side distillation of client models~\cite{lin2020ensemble} has in general been found to outperform simpler aggregation schemes, such as FedAvg.

Recently, \cite{kassab2022federated} proposed an approach for Bayesian federated learning which represents each client's posterior using a set of particles. However, in each round, all the particles needs to be sent to the server, making both communication as well as server-side aggregation very expensive. 

\section{Experiments}
In this section, we compare our client uncertainty-driven probabilistic federated learning approach with various relevant baselines on several benchmark datasets. We report results on the following tasks: (1) Classification in federated setting, (2) Active Learning in federated setting, and (3) OOD detection on each client. In this section, we refer to our approach with simple averaging on the server side as FedPPD, and the variant with distillation based aggregation on the server side as FedPPD+Distill. We have also made our code publicly available at \url{https://github.com/aishgupta/fedppd}.

\subsection{Experimental Setup}
\subsubsection{Baselines} We compare our methods with the following baselines

(1) \textbf{FedAvg}~\citep{mcmahan2017communication} is the standard federated algorithm in which the local models of the participating clients are aggregated at server to compute a global model which is then sent back to all the clients for initialization in the next round.

(2) \textbf{FedBE}~\citep{chen2020fedbe} is another state-of-the-art baseline which provides a more robust aggregation scheme. Instead of only averaging the client models at the server, a probability distribution is fit using the client models, several other models are generated from this probability distribution, and then the client models as well as the generated models are distilled into a single model to yield the global model at the server, which is sent to all the clients for initialization in the next round. Note however that the clients in FedBE only perform point estimation of their weights unlike our approach which estimates the posterior distribution and the PPD of each client. 

(3) \textbf{Federated SWAG}~\citep{linsner2021approaches} is a Bayesian federated learning algorithm which is essentially based on a federated extension of the SWAG~\cite{maddox2019simple} which is an efficient Bayesian inference algorithm for deep neural networks. However, Federated SWAG relies on a simplification that it executes standard federated averaging for all except the last round and in the last round, the SWAG algorithm is invoked at each client to yield a posterior. Also note that Federated SWAG requires Monte-Carlo sampling at test time (thus relying on ensemble based slow prediction) unlike our method which only requires a single neural network to make the prediction.

In addition to the above baselines, in the appendix, we also provide a comparison with \textbf{FedPA}~\citep{al2020federated}, a Bayesian federated learning method, which estimates local posteriors (assumed to be Gaussian) over the client weights, and aggregates them at the server to form an approximate global posterior.

\subsubsection{Datasets}
We evaluate and compare our approach with baseline methods on four datasets: MNIST~\citep{lecun-mnisthandwrittendigit-2010}, FEMNIST~\citep{cohen2017emnist}, and CIFAR-10/100~\citep{krizhevsky2009learning}. MNIST comprises of images of handwritten digits categorized into 10 classes. It has a total of 60,000 images for training and 10,000 images for testing. FEMNIST consists of images of handwritten characters (digits, lowercase, and uppercase alphabets resulting in total of 62 classes) written by multiple users. It has a total of 80,523 images written by 3,550 users. CIFAR-10 consists of $32\times32$ dimensional RGB images categorised into 10 different classes. It has a total of 50,000 images for training and 10,000 images for testing. CIFAR-100 is similar to CIFAR-10 but has 100 distinct classes.

\subsubsection{Model Architecture and Configurations}
\label{sec:config}
In all our experiments, the student model has a larger capacity compared to teacher model as it is modeling the PPD by distilling multiple models drawn from the posterior distribution. We have used a customized CNN architecture for both teacher and student model on MNIST, FEMNIST and CIFAR-10 dataset, with student model being deeper and/or wider than its corresponding teacher model. For CIFAR-100, ResNet-18 and ResNet-34 are used as the teacher and student model, respectively.

In all our experiments, we consider $K=10$ clients with data heterogeneity (experimental results for IID setting are reported in the appendix). Each client holds a small non-i.i.d. subset of training data - approximately 2000 samples for FEMNIST, CIFAR-10 and CIFAR-100 and around 500 samples for MNIST. Except for the FEMNIST data where we have used the Leaf~\citep{caldas2018leaf} benchmark to distribute data among clients (excluding digits to increase class imbalance), for all other datasets clients strictly maintains a small subset of all the classes (at most 2 major classes for MNIST and CIFAR-10 and at most 20 major classes for CIFAR-100)
For a fair comparison, we run our method and all the baselines for 200 rounds on all the datasets (except MNIST where we run it for 100 rounds) and train local client model for 10 epochs in each round. Also, we assume complete client participation i.e. all the clients are considered in each round. However, we tune the learning rate, momentum and weight decay for each method independently. For FedBE and FedPPD$+$Distill, we run additional 20 and 50 epochs at the server for distillation on CIFAR/MNIST and FEMNIST datasets, respectively. 

\begin{figure}[!htb]
    \centering
    \small
    \includegraphics[scale=0.5]{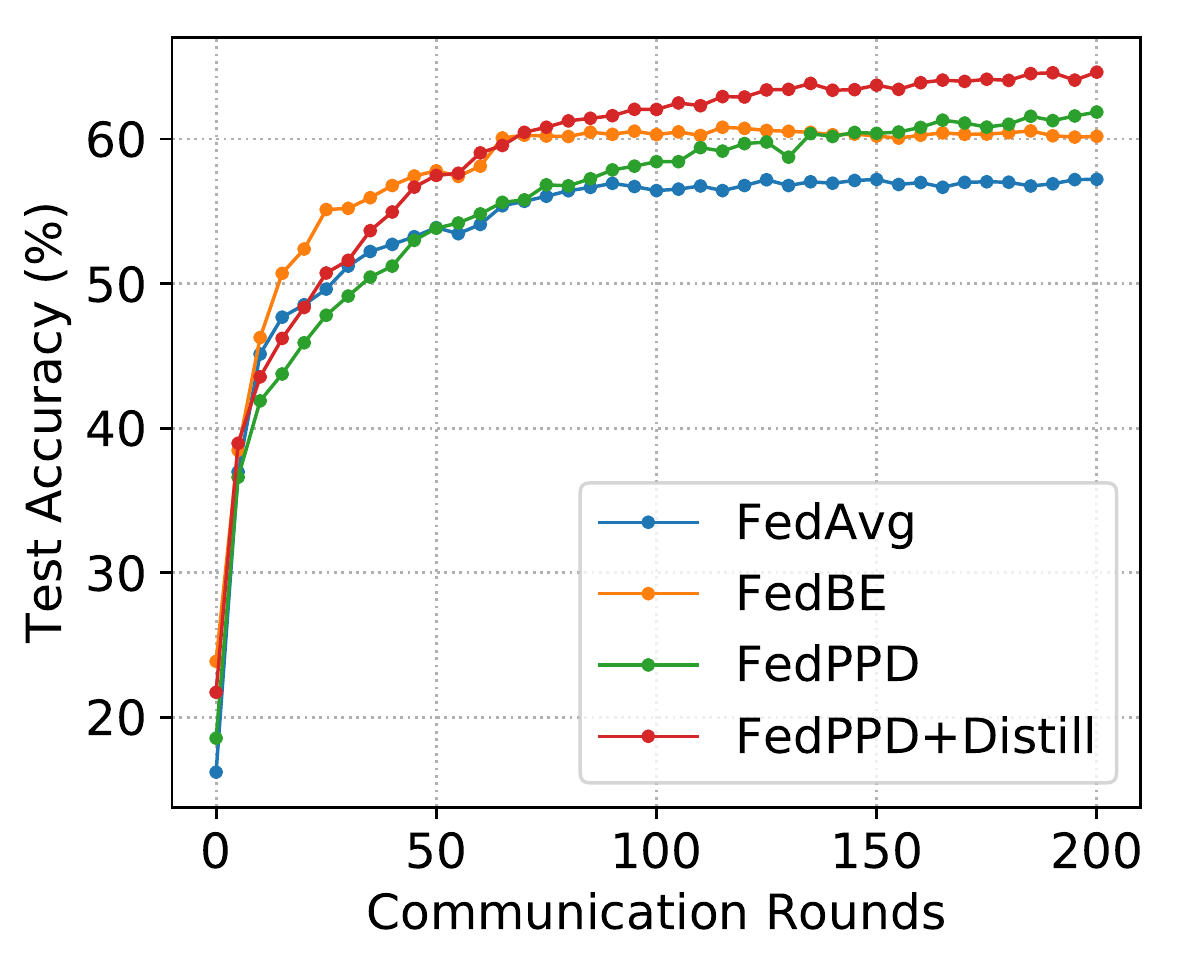}
    \caption{Convergence of all the methods on CIFAR-10 dataset}
    \label{fig:convergence_plot}
    \vspace{-1.5em}
\end{figure}

\subsection{Tasks}
\textbf{Classification} 
We evaluate FedPPD (its two variants) and the baselines on several classification datasets and report the accuracy on the respective test datasets. The results are shown in Table~\ref{tab:classification_acc}. We also show the convergence of all the methods on CIFAR-10 in Figure~\ref{fig:convergence_plot} (similar plots for other datasets are provided in the appendix). Both the variants of FedPPD outperform the other baselines on all the datasets. As compared to the best performing baseline, our approach yields an improvement of $4.44\%$ and $7.08\%$ in accuracy on CIFAR-10 and CIFAR-100, respectively. On MNIST and FEMNIST datasets too, we observe noticeable improvements. The improvements across the board indicate that FedPPD and its variants are able to leverage model uncertainty to yield improved predictions especially when the amount of training data per client is small, which is the case with the experimental settings (as mentioned in Sec~\ref{sec:config}). We also observe that in cases where there is significant heterogeneity in the data distribution across the different clients (on CIFAR-10 and CIFAR-100), the performance gains offered by FedPPD and its variants are much higher as compared to the baselines. On other simpler tasks like MNIST dataset and FEMNIST with the data distribution being roughly similar across different clients, FedPPD results in reasonable gains in the performance. We also quantify the calibration error of our approach and all the baselines models and report the results in Table~\ref{tab:calibration_metrics} where FedPPD and its variant have the least error on the test dataset.

The improved performance of our algorithm FedPPD can be attributed to the following reasons: (1) ability
to incorporate model uncertainty which helps us compute the predictive uncertainty, whereas
other approaches rely on point estimates of the model parameters; (2) although FedSWAG also captures model uncertainty, it uses a crude Gaussian approximation of the
posterior whereas FedPPD do not make any such strict assumptions. 

\begin{table}[!htbp]
    \centering
    \small
    \begin{tabular}{ccccc}
        \toprule
        Model & MNIST & FEMNIST & CIFAR-10 & CIFAR-100 \\
        \midrule
        FedAvg         & 97.74 & 87.40 & 57.20 & 47.02 \\
        FedAvg+SWAG    & 97.75 & 87.45 & 57.34 & 47.07 \\
        FedBE          & 97.82 & 88.12 & 60.18 & 47.52 \\
        FedPPD         & 97.85 & \textbf{88.81} & 61.86 & 53.00 \\
        FedPPD+Distill & \textbf{98.08} & 88.80 & \textbf{64.62} & \textbf{54.60} \\
        \bottomrule
    \end{tabular}
    \caption{Federated classification test accuracies on benchmark datasets}
    \label{tab:classification_acc}
    \vspace{-1em}
 \end{table}  
 
 \begin{table}[!htbp]
    \centering
    \small
    \begin{tabular}{|c|c|c|c|c|c|c|}
    \toprule      & \multicolumn{3}{c|}{CIFAR-10} & \multicolumn{3}{c|}{CIFAR-100}\\
    \midrule
        Model & ECE & MCE & BS  & ECE & MCE & BS \\
        \hline
        FedAvg         & 16.88 & 24.71 & 0.60 & 28.83 & 42.26 & 0.78\\
        FedAvg+SWAG    & 16.69 & 23.16 & 0.60 & 29.07 & 44.36 & 0.78\\
        FedBE          & 19.26 & 27.54 & 0.59 & 31.89 & 45.92 & 0.80\\
        FedPPD         & \textbf{4.31} & \textbf{6.62} & 0.50   & 13.92 & \textbf{21.19} & 0.63\\
        FedPPD+Distill & 10.83 & 16.71 & \textbf{0.49} & \textbf{13.57} & 23.69 & \textbf{0.61}\\
        \bottomrule
    \end{tabular}
    \caption{Expected Calibration Error (ECE), Maximum Calibration Error (MCE) and Brier-score (BS) of all the models on CIFAR-10 and CIFAR-100}
    \label{tab:calibration_metrics}
    \vspace{-1em}
\end{table}

\begin{figure}[h]
    \centering
    \small
    \includegraphics[scale=0.5]{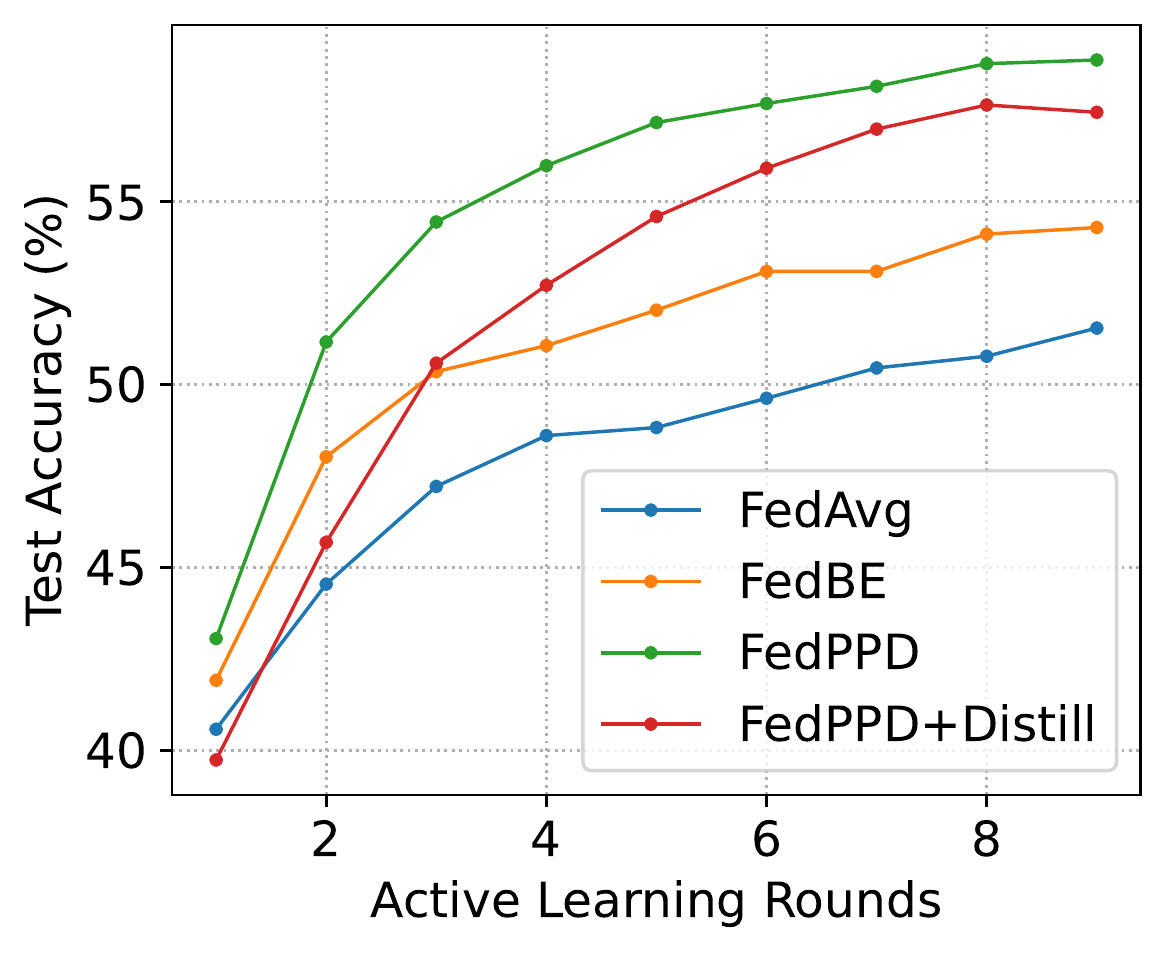}
    \caption{Federated Active Learning on CIFAR-10. Note: FedAvg+SWAG performed almost similarly to FedAvg on this task as well so we skip it from the plot.}
    \label{fig:al_results}
\end{figure}

\textbf{Federated Active Learning} 
In active learning, the goal of the learner is to iteratively request the labels of the most informative inputs instances, add these labeled instances to the training pool, retrain the model, and repeat the process until the labeling budget remains. Following~\citep{ahn2022federated}, we extend our method and the baselines to active learning in federated setting using entropy of the predictive distribution of an input $x$ as the acquisition function. The entropy-based acquisition function for input $x$ is computed as $I(x) = -\sum_{i=1}^{C} p(y=i|x) \log p(y=i|x)$ ($C$ refers to the number of classes) and is used as a score function. In federated active learning setting (we provide a detailed sketch of the federated active learning algorithm in the appendix), each client privately maintains a small amount of labeled data and a large pool of unlabeled examples. In each round of active learning, clients participate in federated learning with their currently labeled pool of data until the global model has converged. Now, each client uses the global model to identify a fixed number (budget) of the most informative inputs among its pool of unlabeled input based on the highest predictive entropies $I(x)$, which are then annotated (locally maintaining data privacy) and added to the pool of labeled examples. Now, with this, next round of active learning begins, where clients will participate in federated learning and use the global model to expand their labeled pool. This process continues until either the unlabeled dataset has been exhausted completely or desired accuracy has been achieved. For a fair comparison, we have run federated active learning on CIFAR-10 dataset with same parameters for all the approaches. We start active learning with 400 labeled and 3200 unlabeled samples at each client and use a budget of 400 samples in every round of active learning. For federated learning, we use the same hyperparameters as for the classification experiments. We stop federated active learning once all the clients have exhausted their unlabeled dataset and show the results in Figure~\ref{fig:al_results} where FedPPD and its variant attain the best accuracies among all the methods.

\textbf{Out-of-distribution (OOD) detection} We also evaluate FedPPD and its variant, and the other baselines, in terms of their ability to distinguish between Out-of-Distribution (OOD) data and data used during training phase (in-distribution data). 
For this, given any sample $x$ to be classified among $k$ distinct classes and model weights $\theta$ (or PPD for our approach), we compute Shannon entropy of the model's predictive distribution for the input $x$ and compute the AUROC (Area Under ROC curve) metric. 
We use KMNIST as OOD data for models trained on FEMNIST, and SVHN for CIFAR-10/CIFAR-100 models. Note that, to avoid class imbalance, we sample an equal amount of data for both the distributions (out and in) and repeat it 5 times. We report the results in Table~\ref{tab:auroc_OOD} where FedPPD and its variant consistently result in better AUROC scores on all the datasets validating its robustness and accurate estimates of model uncertainty. In addition to OOD detection, we also apply all the methods for the task of identifying the correct predictions and incorrect predictions based on the predictive entropies. For this task too, FedPPD and its variant outperform the other baselines as shown in Table~\ref{tab:auroc_correct}. 

\begin{table}[!htbp]
    \centering
    \setlength\tabcolsep{3pt}
    \small
    \begin{tabular}{|c|c|c|c|}
        \toprule
        Model & FEMNIST & CIFAR-10 & CIFAR-100 \\
        \midrule
        FedAvg         & $0.957 \pm 0.003$ & $0.728 \pm 0.013$  & $0.703 \pm 0.011$ \\
        FedAvg+SWAG    & $0.956 \pm 0.003$ & $0.728 \pm 0.013$  & $0.704 \pm 0.011$\\
        FedBE          & $0.966 \pm 0.003$ & $0.728 \pm 0.006$  & $0.669 \pm 0.009$ \\
        FedPPD         & $\mathbf{0.983 \pm 0.003}$ & $0.701 \pm 0.007$  & $0.698 \pm 0.009$ \\
        FedPPD+Distill & $0.949 \pm 0.003$ & $\mathbf{0.765 \pm 0.006}$  &  $\mathbf{0.784 \pm 0.008}$ \\
        \bottomrule
    \end{tabular}
    \caption{AUROC score for OOD/in-domain data detection}
    \label{tab:auroc_OOD}
    \vspace{-1.5em}
\end{table}

\begin{table}[!htbp]
    \centering
    \setlength\tabcolsep{3pt}
    \small
    \begin{tabular}{|c|c|c|c|}
        \toprule
        Model & FEMNIST & CIFAR-10 & CIFAR-100 \\
        \midrule
        FedAvg         & $0.846 \pm 0.011$ & $0.742 \pm 0.011$  & $0.792 \pm 0.003$ \\
        FedAvg+SWAG    & $0.845 \pm 0.009$ & $0.743 \pm 0.010$  & $0.800 \pm 0.004$\\
        FedBE          & $\mathbf{0.863 \pm 0.005}$ & $0.753 \pm 0.007$  & $0.789 \pm 0.005$\\
        FedPPD         & $0.862 \pm 0.008$ & $0.755 \pm 0.007$  & $0.814 \pm 0.003$\\
        FedPPD+Distill & $0.853 \pm 0.013$ & $\mathbf{0.769 \pm 0.006}$  &  $\mathbf{0.823 \pm 0.002}$\\
        \bottomrule
    \end{tabular}
    \caption{AUROC score for correct/incorrect data detection}
    \label{tab:auroc_correct}
    \vspace{-1.5em}
\end{table}

\section{Conclusion and Discussion}
Leveraging model uncertainty in federated learning has several benefits as we demonstrate in this work. To achieve this, we developed a uncertainty-driven approach to federated learning by leveraging the idea of distilling the posterior predictive into a single deep neural network.
In this work, we consider a specific scheme to distill the PPD at each client. However, other methods that can distill the posterior distribution into a single neural network~\citep{wang2018adversarial,vadera2020generalized} are also worth leveraging for probabilistic federated learning. Another interesting future work will be to extend our approach to settings where different clients could possibly be having different model architectures. Finally, our approach first generates MCMC samples (using SGLD) and then uses these samples to obtain the PPD in form of a single deep neural network. Recent work has shown that it is possible to distill an ensemble into a single model without explicitly generating samples from the distribution~\citep{ratzlaff2019hypergan}. Using these ideas for uncertainty-driven probabilistic federated learning would also be an interesting future work.  

\acks{AG acknowledges the Prime Minister's Research Fellowship (PMRF) for the support. PR acknowledges support from Google Research India.}

\small
\bibliography{references}

\appendix

\section{Learning Posterior Predictive Distribution (PPD) locally at each client}\label{apd:first}

At each client, we aim to distill the Monte Carlo approximation of its PPD into a single neural network. We do this using an online approach, similar to~\citep{korattikara2015bayesian}. We maintain two deep neural networks on each client: 1) the first neural network is optimized using SGLD on client's private dataset to draw samples from its posterior (these samples denote the collection of teacher models on each client), and 2) the second neural network represents the distilled version of the client's Monte Carlo approximation of its PPD (student model). In each local iteration at client, we draw a sample (i.e., a teacher model) from its posterior distribution and  distill it into the student model which represents the PPD. This incremental distillation process, when all the teacher models are distilled into the student model, ultimately gives the student model representing the PPD (in form of a single deep neural network) at the client. We summarize the client updates in Algorithm~\ref{alg:bdkalg}.

\begin{algorithm2e}[h]
\DontPrintSemicolon
\SetNoFillComment
\caption{Local update at client $k$}\label{alg:bdkalg}
\KwData{Dataset $\mathcal{D}_k = {(x_i, y_i)}_{i=1}^{N_k}$, batch-size $M_k$, number of local iterations $V$, teacher model weights $\theta$, student model weights $w$, teacher learning rate at iteration $v$, $\alpha^v$, student learning rate at iteration $v$, $\beta^v$, teacher prior hyperparameters $\gamma_k$, student prior hyperparameters $\mu_k$}
\BlankLine
\For{\texttt{$v$ = 0 ... $V-1$}}{
    \tcc{teacher model update}
    Sample a minibatch $B$ of size $M_k$\;
    Sample Gaussian Noise $z_v \sim \mathcal{N}(0, \alpha^v I)$ \;
    \BlankLine
    $\theta_k^{v+1} = \theta_k^v + \frac{\alpha^v}{2} \nabla_{\theta_k}(\log p(\theta_k^v | \gamma_k) + \frac{N_k}{M_k} \sum_{(x,y)\in \mathcal{B}} \log p(y | x, \theta_k^v)) + z_v$ \;
    Generate minibatch $B^\prime$ of size $M_k$ \;
    \BlankLine
    \tcc{student model update}
    Add Gaussian Noise to $B^\prime$ \;
    $w_k^{v+1} = w_k^v + \beta^v \nabla_{w_k} (\frac{1}{M_k} \sum_{x \in B^\prime} \sum_c p(y = c | x, \theta_k) \log p(y = c | x, w_k^v) + \log p(w_k^v | \mu_k))$.\;
}
\end{algorithm2e}

\section{Federated Active Learning}\label{apd:second}

Active learning is an iterative process which aids a learner in achieving desired performance with limited number of labeled input instances. In each iteration of active learning, the learner identifies the most informative inputs, requests their labels, adds them to its pool of labeled instances and retrain the model with augmented labeled dataset. This process repeats until the labeling budget is not fully exhausted. Similarly, in federated active learning~\citep{ahn2022federated}, active learning can be performed on each client using the global model (informative of global data distribution) instead of its local model to identify the most helpful instances. Thus, in each iteration of federated active learning, each client identifies the most helpful inputs, annotates it locally (or using oracle preserving data privacy), adds it to its local dataset and participates in federated learning until the convergence of the global model. In our work, we use entropy of the model's output as the score function to identify the most helpful inputs, though other predictive uncertainty based score functions used in active learning can be employed as well. Also, note that to compute $p(y|x)$, FedPPD uses posterior predictive distribution whereas the other baseline methods like FedAvg uses the point-estimate of the global model. Our federated active learning algorithm is sketched in detail in Algorithm~\ref{alg:fal}.

\begin{algorithm2e}
\DontPrintSemicolon
\BlankLine
\caption{Federated Active Learning}\label{alg:fal}
\KwData{Number of active rounds $A$, client id $\{1, 2, 3, \dots, k\}$, client labeled dataset $\{\mathcal{L}_i\}_{i=1}^k$, client unlabeled dataset $\{\mathcal{U}_i\}_{i=1}^k$, budget per round $\mathcal{B}$, set of inputs to be annotated $S'$, global server model $\theta$}
\For{each round $r = 0, \dots, A-1$}{
    \For{each client $i = 0, \dots, k$}{
        $S_i = \{\}$ \;
        \For{each input $x \in \mathcal{U}$}{
            $p(y|x) = \theta_r(x)$ \;
            $I(x) = -\sum_{i=1}^{C} p(y=y_i|x) \log p(y=y_i|x)$ \;
            $S_i=S_i\cup \{(x, I(x)\}$ \;
        }
        Select subset $S_i^\prime$ of size $\mathcal{B}$ from $S_i$ with maximum entropy and get it annotated \;
        $\mathcal{L}_i = \mathcal{L}_i \cup S_i^\prime \qquad \mathcal{U}_i = \mathcal{U}_i-S_i^\prime$ \;
    }   
    $\theta_{r+1} = $ Updated global model using federated learning on $\{\mathcal{L}_i\}_{i=1}^k$ \;
}
\end{algorithm2e}

\section{Experimental Setup}\label{apd:third}

We now provide the implementation details of the experiments mentioned in the main paper. 

\textbf{Model architecture}
We evaluate all the variants of FedPPD and baseline algorithms on MNIST, FEMNIST, and CIFAR-10 using customized CNNs and use ResNets for CIFAR-100. To have a fair comparison, the architecture of the teacher model in FedPPD and client model in all the baseline approaches is the same. We provide the architecture details of the teacher and student model for all datasets in Table~\ref{tab:model_arch_teach} and \ref{tab:model_arch_stud}, respectively.

\begin{table}[!htbp]
    \centering
    \small
    \begin{tabular}{|c|c|c|}
    \toprule
    MNIST & FEMNIST & CIFAR-10 \\
    \midrule
    Conv2D$(5 \times 5, 10)$ & Conv2D$(5 \times 5, 32)$  & Conv2D$(5 \times 5, 6)$ \\
    MaxPool$(2 \times 2)$ & MaxPool$(2 \times 2)$ & MaxPool$(2 \times 2)$    \\
    Conv2D$(5 \times 5, 20)$ &  Conv2D$(5 \times 5, 64)$ & Conv2D$(5 \times 5, 16)$ \\
    Dropout2D$(0.5)$ & Dropout2D$(0.5)$ & MaxPool$(2 \times 2)$\\
    MaxPool$(2 \times 2)$ & MaxPool$(2 \times 2)$ & Linear$(120)$\\
    Linear$(50)$ & Linear$(128)$ & Linear$(84)$\\
    Dropout$(0.5)$ & Dropout$(0.5)$ & Linear$(10)$\\
    Linear$(10)$ & Linear$(52)$ & \\
    \bottomrule
    \end{tabular}
    \caption{Model Architecture for Teacher Network and Baselines Methods}
    \label{tab:model_arch_teach}
\end{table}

\begin{table}[!htbp]
    \centering
    \small
    \begin{tabular}{|c|c|c|}
    \toprule
    MNIST & FEMNIST & CIFAR-10 \\
    \midrule
    Conv2D$(5 \times 5, 20)$ & Conv2D$(5 \times 5, 50)$  & Conv2D$(5 \times 5, 16)$ \\
    MaxPool$(2 \times 2)$ & MaxPool$(2 \times 2)$ & MaxPool$(2 \times 2)$    \\
    Conv2D$(5 \times 5, 40)$ &  Conv2D$(5 \times 5, 100)$ & Conv2D$(5 \times 5, 16)$ \\
    Dropout2D$(0.5)$ & Dropout2D$(0.5)$ & MaxPool$(2 \times 2)$\\
    MaxPool$(2 \times 2)$ & MaxPool$(2 \times 2)$ &  Linear$(256)$\\
    Linear$(100)$ & Conv2D$(5 \times 5, 200)$ & Linear$(128)$\\
    Dropout$(0.5)$ & Dropout2D$(0.5)$ & Linear$(10)$\\
    Linear$(10)$ & MaxPool$(2 \times 2)$ & \\
    & Linear$(1600)$ & \\
    & Dropout$(0.5)$ & \\
    & Linear$(180)$ & \\
    & Dropout$(0.5)$ & \\
    & Linear$(52)$ & \\
    \bottomrule
    \end{tabular}
    \caption{Model Architecture for Student Networks}
    \label{tab:model_arch_stud}
\end{table}

\textbf{Hyperparameters}
We tune the hyperparameters (learning rate and weight decay) on each dataset for all the variants of FedPPD and baseline methods. The optimal learning rate of the teacher and student model during local learning in FedPPD are $\{0.045, 0.055\}$, $\{0.050, 0.085\}$ and $\{0.055, 0.020\}$ on MNIST, FEMNIST and CIFAR respectively. In case of FedPPD with distillation at server, the teacher and student learning rates during local training are $\{0.045, 0.055\}$, $\{0.060, 0.085\}$ and $\{0.055, 0.020\}$ on MNIST, FEMNIST and CIFAR respectively. Also, during distillation at server, teacher and student model are updated using SWA optimizer with learning rates of $\{0.0010, 0.0010\}$ and $\{0.0015, 0.0025\}$ for MNIST/FEMNIST and CIFAR-10/100 respectively.

\begin{figure*}[h]
    \centering
    \begin{tabular}{c c}
         \subfigure[FEMNIST]{\includegraphics[scale=0.6]{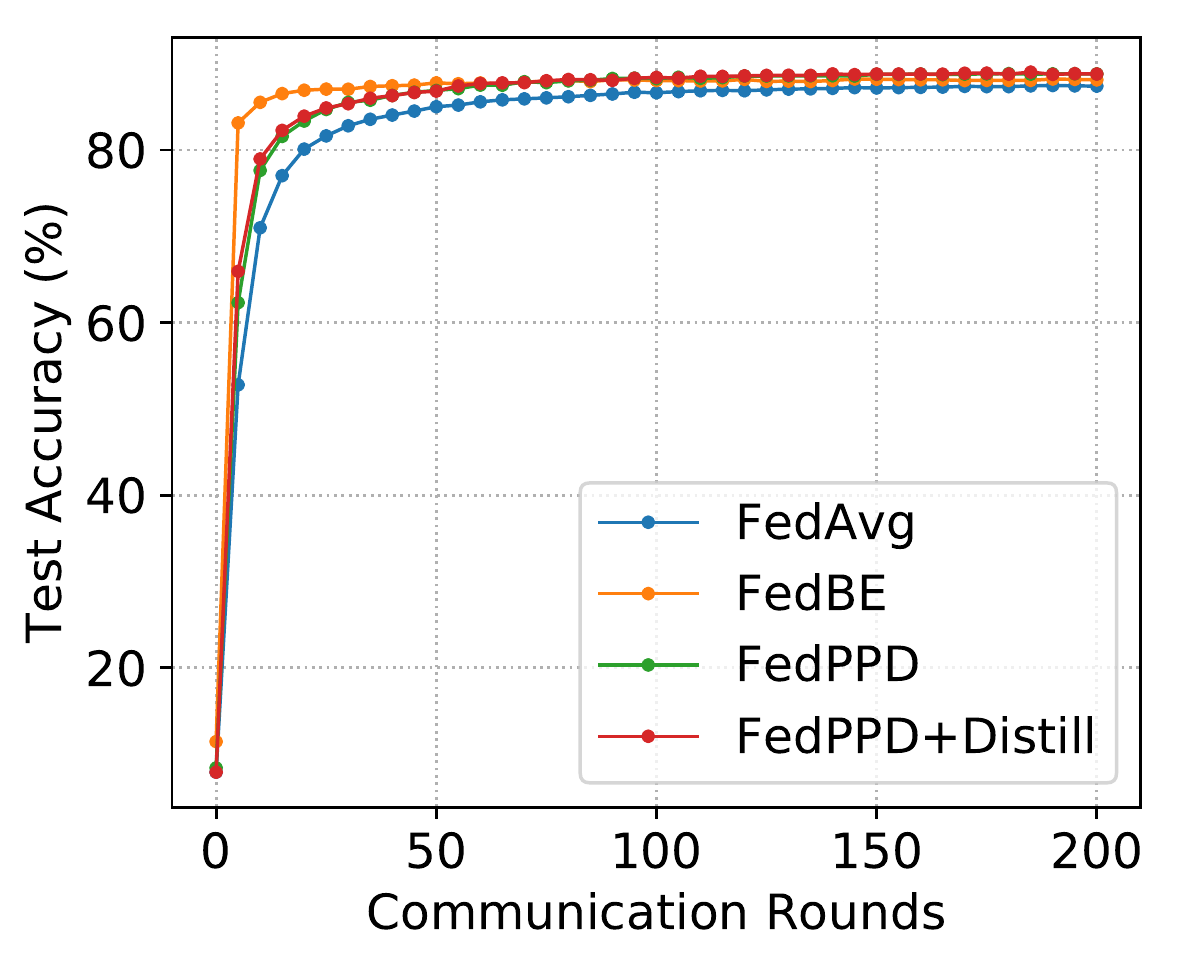}}
         &   
         \subfigure[CIFAR-100]{\includegraphics[scale=0.6]{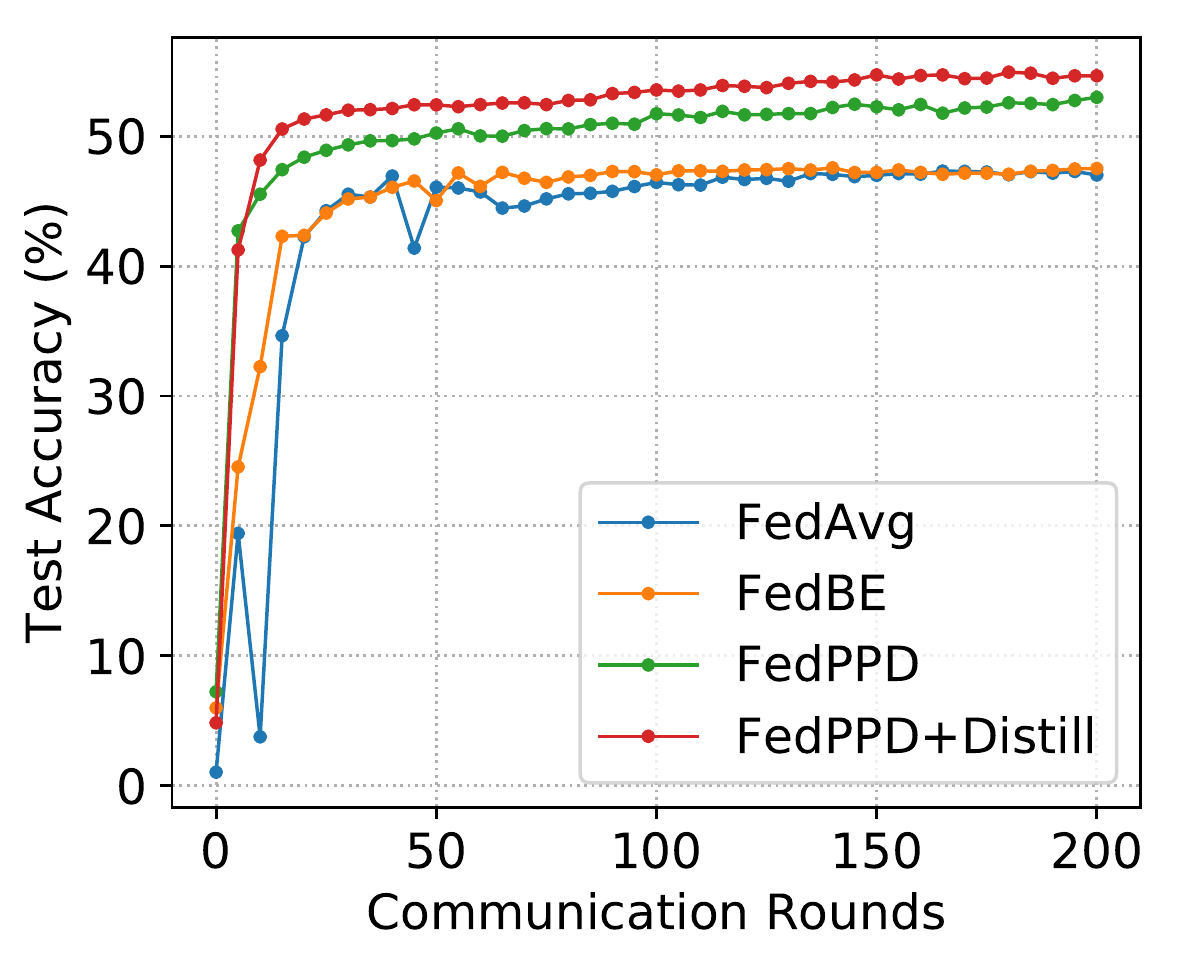}}
    \end{tabular}
    \caption{Convergence of different federated approaches}
    \label{fig:convergence}
\end{figure*}

\textbf{Baseline}
We compare our approach with FedAvg~\citep{mcmahan2017communication} and FedBE~\citep{chen2020fedbe} on MNIST, FEMNIST, CIFAR-10 and CIFAR-100 and the results are presented in the main paper. Here, we provide the model convergence plot for all the approaches on FEMNIST and CIFAR-100 in Fig~\ref{fig:convergence}, showing the superior performance of FedPPD and its variants as compared to the other baselines.

We also compare our method against FedPA~\citep{al2020federated} (approximates the posterior distribution by a Gaussian) using their publicly available code obtained from  ( \url{https://github.com/google-research/federated/tree/master/posterior_averaging}). We extended their implementation to our experimental setting as described in main paper. Unfortunately, the results of FedPA in our experimental setup are either comparable or worse than FedAvg. We have reported the results of FedPA on all the datasets in Table~\ref{tab:fedpa_acc}. A possible reason for the poor performance of FedPA could be because the Gaussian approximation of global posterior may not be accurate enough, and moreover since, at the server, FedPA only computes the Gaussian posterior's mode/mean but not the covariance, using only the mode/mean ignores the uncertainty, leading to poorer predictions. 

\begin{table}[h]
    \centering
    \setlength\tabcolsep{4pt}
    \begin{tabular}{c c c c c}
        \toprule
            & MNIST & FEMNIST & CIFAR-10 & CIFAR-100 \\
        \midrule
        FedPA & 96.86 & 86.47 & 50.88 & 42.18 \\
    \bottomrule
    \end{tabular}
    \caption{Performance of FedPA on test dataset}
    \label{tab:fedpa_acc}
\end{table}

\textbf{IID Data Distribution}
We now consider a setting in which the data is distributed among clients in IID fashion. We have experimented on CIFAR-10 and CIFAR-100 dataset and compared our proposed approach with the baseline methods. We have assigned a subset of 4000 randomly selected labeled images to 10 client and have maintained a small subset of 10000 unlabeled images as the proxy dataset on the server for FedBE and FedPPD$+$Distill. We have also considered complete client participation i.e. all the 10 clients participate in every communication round of the federated learning. Also, for all the remaining hyperparameters, we have used same values as we did for the non-IID setting. The results of the experiments are reported in Table~\ref{tab:iid_dist} where FedPPD and its variant clearly outperforms the baseline methods. This shows that FedPPD and FedPPD$+$Distill result in improved performance even when the data distribution among clients is homogeneous.

\begin{table}[h]
    \centering
    \small
    \setlength\tabcolsep{4pt}
    \begin{tabular}{c c c c c}
        \toprule
            & FedAvg & FedBE & FedPPD & FedPPD+Distill \\
        \midrule
        CIFAR-10  & 73.80\% & 73.99\% & 74.91\% & \textbf{75.26\%} \\
        CIFAR-100 & 53.38\% & 53.16\% & \textbf{63.67}\% & 63.43\% \\
    \bottomrule
    \end{tabular}
    \caption{Classification performance of the global model on the test dataset for IID data distribution among clients}
    \label{tab:iid_dist}
\end{table}

\textbf{Resources used}
We ran all our experiments on Nvidia 1080 Ti GPUs with 12 GB of memory. We have implemented our method in PyTorch and utilized its multiprocessing  library to spawn multiple threads for parallel computation.

\section{Potential Limitations/Future Work}\label{apd:fourth}

We have shown the efficacy and robustness of our approach against multiple baseline methods on various datasets. Even though, to obtain (an approximation to) the PPD at each client, we use a specific approach based on stochastic gradient MCMC (SGMCMC) combined with knowledge distillation, our work provides a general framework for Bayesian federated learning where we can use a variety of methods at each client to obtain the posterior/PPD approximation, and then leverage techniques developed for standard federated learning. The key is to represent the PPD approximation via a single deep neural network. In our work, we use the vanilla SGMCMC at the clients which sometimes can have convergence issues. However, recent process on SGMCMC has led to more robust variants of SGMCMC which can be employed under our framework for better performance.

Generation of posterior samples using MCMC/SGMCMC and then distilling them into a student model (even using an online procedure like us) can be slow, especially since MCMC methods can sometimes exhibit slow convergence. One possible avenue of future work could be to represent the posterior of the model implicitly and distill it without having to explicitly generate samples from it~\citep{ratzlaff2019hypergan}.
\end{document}